\documentclass[runningheads]{llncs}

\usepackage{eccv}

\usepackage{eccvabbrv}

\usepackage{graphicx}
\usepackage{booktabs}

\usepackage[accsupp]{axessibility}  % Improves PDF readability for those with disabilities.

\usepackage[pagebackref,breaklinks,colorlinks,citecolor=eccvblue]{hyperref}
\usepackage{orcidlink}

\usepackage{makecell}
\usepackage{amsmath}
\usepackage{amssymb}

\begin{document}

\title{ViLA: Efficient Video-Language Alignment for Video Question Answering} 

\titlerunning{ViLA}

\author{Xijun Wang\inst{1} \and
Junbang Liang\inst{2} \and
Chun-Kai Wang\inst{2}\and
Kenan Deng\inst{2}\and
Yu Lou\inst{2}\and
Ming C. Lin\inst{1,2}\and
Shan Yang\inst{2}
}

\authorrunning{Xijun Wang et al.}

\institute{University of Maryland, College Park, USA \and
Amazon, USA \\
\email{\{xijun, lin\}@umd.edu, \{junbangl, kenanden, ckwang, ylou, minglinz, ssyang\}@amazon.com}}

\maketitle

\begin{abstract}
We propose an efficient \textbf{Vi}deo-\textbf{L}anguage 
\textbf{A}lignment (ViLA) network.
Our ViLA model addresses both efficient frame sampling and effective cross-modal alignment in a unified way. 
In our ViLA network, we design a new learnable text-guided Frame-Prompter together with a cross-modal distillation (QFormer-Distiller) module.
Pre-trained large image-language models have shown promising results on problems such as visual question answering (VQA).
However,
how to efficiently and effectively sample video frames when 
adapting pre-trained large image-language model to video-language alignment is still the major challenge.
Compared with prior work, our ViLA model demonstrates the capability of selecting key frames with critical contents, thus improving the video-language alignment accuracy while reducing the inference latency ({\bf +3.3\%} on NExT-QA Temporal with {\bf3.0$\times$} speed up). 
Overall, our ViLA network outperforms the state-of-the-art methods on the video question-answering benchmarks:
{\bf+4.6\%} on STAR Interaction, {\bf+2.2\%} on STAR average with {\bf3.0$\times$} speed up, ours 2-frames out-perform SeViLA 4-frames on the VLEP dataset with {\bf 4.2$\times$} speed-up. Code will be available at https://github.com/xijun-cs/ViLA.

\end{abstract}

%In this work, we propose an efficient Video-Language Alignment (ViLA) network. Our ViLA model addresses both efficient frame sampling and effective cross-modal alignment in a unified way. In our ViLA network, we design a new learnable text-guided Frame-Prompter together with a new cross-modal distillation (QFormer-Distiller) module. Pre-trained large image-language models have shown promising results on problems such as visual question answering (VQA). However, how to efficiently and effectively sample video frames when adapting pre-trained large image-language model to video-language alignment is still the major challenge. Compared with prior work, our ViLA model demonstrates the capability of selecting key frames with critical contents, thus improving the video-language alignment accuracy while reducing the inference latency +3.3% on NExT-QA Temporal with 3.0X speed up).  Overall, our ViLA network outperforms the state-of-the-art methods on the video question-answering benchmarks: +4.6% on STAR Interaction, +2.2% on STAR average with 3.0X speed up, ours 2-frames out-perform SeViLA 4-frames on the VLEP dataset with 4.2X speed-up. 
\section{Introduction}
\label{sec:intro}

% To insert a figure: \input{figs/template}
% Or table: \input{tables/template}
“If a picture is worth thousands of words, what is a video worth?”~\cite{introquote} Video watching has become a new social norm. Statistics show YouTube has approximately $122$ million daily active users, based all over the world. Visitors spend on average $19$ minutes per day on YouTube. An average of close to 1 million hours of video are streamed by YouTube users each and every minute. 
As video data continue to grow through internet viewing, video information retrieval becomes more demanding. 
Video data has tremendous capacity to store a vast variety of useful information. 
Compared to image question answering (Q\&A) problem, video QA is more challenging due to one extra temporal dimension.
How to efficiently sample relevant frames from a video with the computing resource constraint remains a long-standing problem in video QA research.
%To enable text conditioned video information retrieval, cross-modality video understanding is needed. 
%Video-text cross-modal alignment is another major challenge. Especially with the recent advancement in LLMs, 
%how to best leverage pre-trained LLM is yet to be studied.

% Use figure* for multi-column figure
\begin{figure}[tp]
\vspace{-0.5em}
    \centering
    \includegraphics[width=0.93\linewidth]{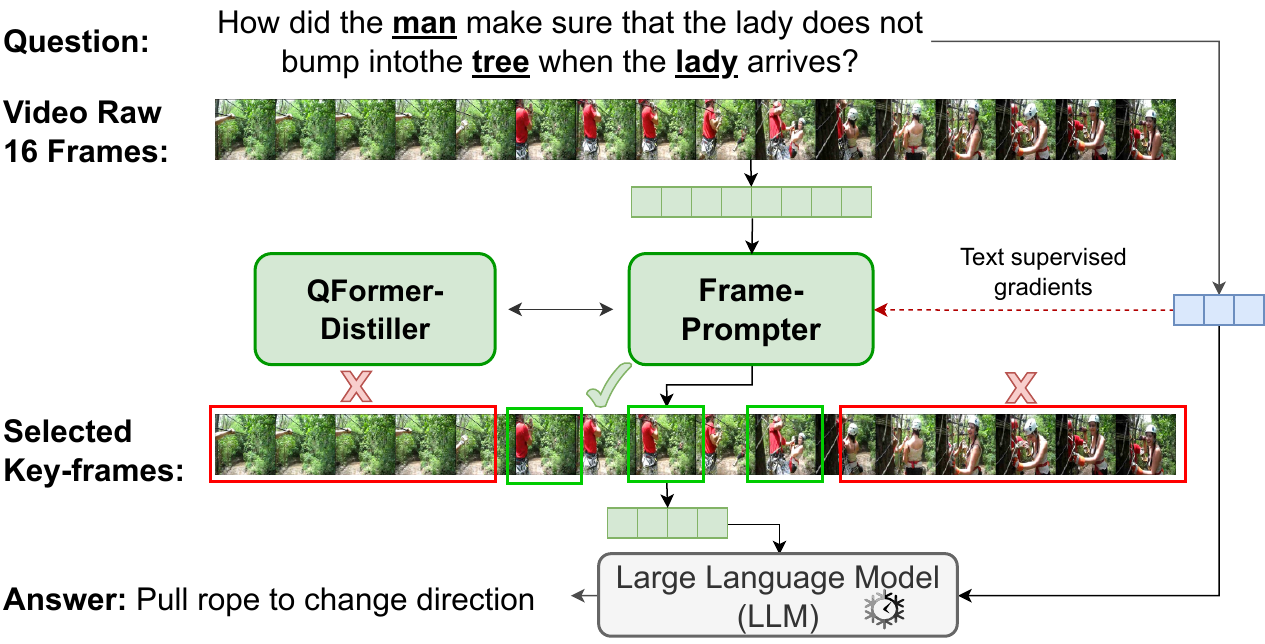}
    \vspace{-0.5em}
    \caption{Our efficient \textbf{Vi}sion-\textbf{L}anguage \textbf{A}lignment (\textbf{ViLA}) model via Frame-Promper and distilling contains two new modules: a {\em text-guided} Frame-Prompter and a cross-modal QFormer-Distiller. It learns to extract the most question-related frames while keeping the inference latency low.
    %\TODO{Update blue embedding to Frame Prompter to red, add text text-supervised}
    }
    \label{fig:task_overview}
\vspace{-2em}
\end{figure}

Recent advances in pre-trained large-scale language models~\cite{brown2020language,touvron2302llama} have greatly boosted the performance of the vision-language models, especially on cross-modality tasks.
Many state-of-the-art image-language models~\cite{chen2022pali, alayrac2022flamingo,li2023blip, wang2023cogvlm} leverage pre-trained LLMs.
These models ~\cite{dai2023instructblip, yang2023mm} achieve excellent performance on visual-language tasks such as image captioning~\cite{dai2023instructblip}, visual question answering~\cite{yang2023mm} and more. 
Inherently, many video-language models~\cite{yu2023self, liu2023improved} are built from these pre-trained image-language models.
These image-based video-language models treat a video as a series of multi-channel images sampled randomly or uniformly~\cite{li2023blip}. 
While this strategy works well for short videos,
for long videos or videos with non-uniform information distribution, random or uniform frame sampling may miss critical information. 
When it comes to video-language alignment, frame sampling efficiency and effectiveness go hand-in-hand.
One needs to not only reduce the number of sampled frames but also select frames that are most related to the input question.
Previous work such as SeViLA\cite{yu2023self} trains a separate keyframe localizer, which is not friendly for the real-time inference and introduces more parameters to the whole model.
%~\ref{tab:nextqa}% 
%The challenge for sampling frames, information localization and efficient training remains even with bootstrapping from large pre-train image-language models.
%\KD{video encoding has a another special meaning. Here it should be frame sampling right?}
Besides video representation, cross-modality alignment while leveraging LLMs is another challenge.
%Cross visual-language alignment has also gained huge improvement with Large-language models. 
%However how to leverage pre-trained powerful LLMs for visual-language is the new challenge.
The critical problem lies in how to efficiently transfer video information to the LLM's input domain. 
%\TODO{here we need to point the contribution of the QFormer-Distiller, any previous work that's only a glue layer, to highlight the Distiller contribution}

% Use figure* for multi-column figure
\begin{figure*}[tp]
\vspace{-2mm}
    \centering
    \includegraphics[width=\linewidth]{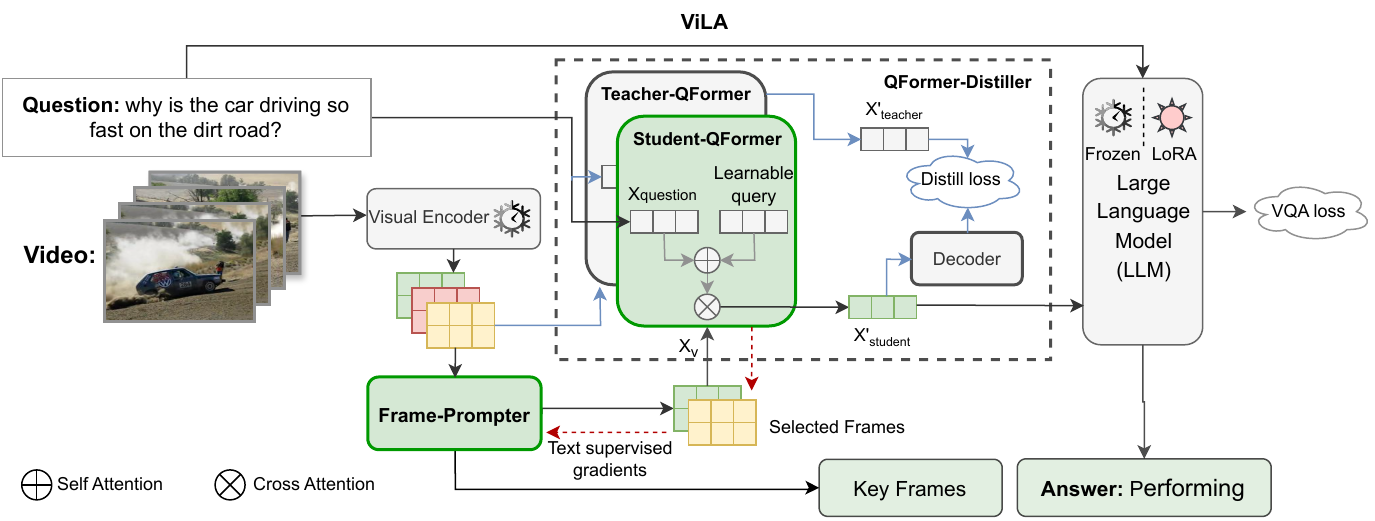}
    \vspace{-5mm}
    \caption{ \textbf{Model Overview.} Our ViLA model includes 4 sub-modules: the visual encoder, text-supervised Frame-Prompter (FP), QFormer-Distiller (QFD), and a LLM. We encode the video frames through a frozen visual encoder. Then we train the Teacher-QFormer using all the frame features.
    After that, we train the Student-QFormer and Frame-Prompter end-to-end. 
    Unlike the Teacher-QFormer, our Student-QFormer is trained with masked frames features from a text-supervised Frame-Prompter.
    Finally, the input question text and QFormer transformed visual features go through a frozen large language model to generate the answer. Our network supports both leveraging LLM through proper visual prompting without affecting the original LLM (Frozen) ability on language tasks and finetuning LLMs(LoRA) simultaneously to get optimal performance on specific tasks.
    %\todo{add LoRA to this figure, and update the caption}
    }
    \label{fig:template}
    \vspace{-4mm}
\end{figure*}
\noindent
{\bf Main Contributions:  } To address these challenges, we propose a new network, ViLA.
%Our ViLA model tackles the problem of both efficient and effective video-language alignment leveraging LLM. 
Compared to the state-of-the-art video-language models~\cite{yu2023self,dai2023instructblip,li2023blip}, ViLA consists of a new Frame-Prompter together with a QFormer-Distiller.
Our Frame-Prompter learns to {\em select the most important frames influenced by the corresponding question text and and supervised by the VQA loss}. Meanwhile, the Frame-Prompter is meticulously designed to keep lightweight so as to be efficient.
%\YS{second thought, not really supervised, loss is supervised, but question text is only influenced}
%Our QFormer-Distiller further reinforces the critical frames selection.
%To align image and language modalities, w
To effectively and efficiently transfer video information to LLM input domain, we add a new distillation on top of the QFormer, named QFormer-Distiller.
The QFormer is the cross-modal query-visual Transformer proposed in previous BLIP models~\cite{dai2023instructblip,li2023blip,li2022blip} for cross-modal fusion.
%We apply the cross-modal query-visual Transformer (QFormer) proposed in previous BLIP models~\cite{dai2023instructblip,li2023blip,li2022blip} for cross-modal fusion.
%To improve the efficiency, we add a new distillation on top of the QFormer.
We train our Frame-Prompter and QFormer-Distiller end-to-end. 
The cross-modal temporal distiller {\em teaches} a smaller (i.e. fewer frames) QFormer.
To the best of our knowledge, this work is the first to propose a Frame-Prompter and a distiller on top of the cross-modal alignment for video-language learning with pre-trained LLMs.
%like BLIP-2~\cite{li2023blip} transfer visual information using a pre-trained visual-to-text module. 
%More recently, in InstructBLIP~\cite{dai2023instructblip}, authors propose a QFormer to fuse and transfer the visual information.

We validate our ViLA model on the Video Question Answering benchmark datasets. 
This includes the NExT-QA~\cite{xiao2021next}, STAR\cite{wu2021star}, How2QA~\cite{li2020hero}, TVQA~\cite{lei2018tvqa} and VLEP~\cite{lei2020more}.
Our work outperforms previous strong SOTA methods
%SeViLA network by on average $2\%$ 
across all the benchmarks, 
while reducing inference latency.
Comparing with SOTA video-language model SeViLA~\cite{yu2023self}, we reduce significant amount ($50\%$) of training parameters and inference latency ($1.35-4.2\times$ speed up), while improving the accuracy.
We also performed ablation study on our text-guided Frame-Prompter and QFormer-Distiller. As shown in Table~\ref{tab:comp}, both the text-guided Frame-Prompter and the QFormer-Distiller play critical roles in making our method effective. 
%by $50\%$.
%Comparing with BLIP-2~\cite{li2023blip}, our new Frame-Prompter and QFormer-Distiller improved the accuracy by on average $5\%$.  

To sum up, the key novelty is a new text-guided Frame-Prompter and question-relevant QFormer-Distiller (trained from end-to-end). The Frame-Prompter enhances the {\em efficiency}, while the later bolsters the {\em effectiveness}. Together they optimize the selection of frames for video-language alignment learning. 
%Our main contributions include:
%\begin{itemize}
%    \item a new {\bf text-guided video frame prompter} to smartly sample important frames together with a cross-modal temporal distillation for efficient and effective temporal learning;
%    \item a {\bf 3-way fusion strategy} to best align vision and language leveraging pre-trained LLM;
%    \item perform experiments on our strategy that out-performs state-of-the-art method on video question answering and captioning benchmarks. 
%\end{itemize}

\section{Related Work}
\label{sec:related}
\subsection{Visual-Language Alignment}
\paragraph{Visual-Language Pre-training}
Vision-Language cross-modal pre-training has greatly improved over the past couple of years.
Various network architectures and pre-training objectives have been proposed for different downstream tasks, including the dual-encoder architecture with image-text contrastive learning~\cite{radford2021learning}, the fusion-encoder architecture with image-text matching~\cite{li2021align}, and unified transformer architecture with masked language modeling~\cite{wang2022image}.
These methods, along with others, focus on the ability to find image-text affinity~\cite{yao2021filip}, correlation~\cite{bao2022vlmo}, and/or completion~\cite{yu2205coca}, and need to pre-train the model end-to-end.
To address the incompatibility with pre-trained unimodal models such as LLMs~\cite{brown2020language}, recent works~\cite{li2023blip} proposed to train a QFormer to bridge the domain gap between two frozen pre-trained models.
Inspired by its flexibility, more downstream tasks and applications have been proposed, including instruction-based image generation~\cite{wu2023visual} and image question-answering~\cite{yang2023mm}.

While most of the previous work focus on image-text alignment, very few have discussed the extension to videos until most recently when temporal modeling starts to be included for better reasoning capabilities.
HiTeA~\cite{ye2022hitea} jointly trains pairs in long and short view to capture temporal relations between moments and event.
Smaug~\cite{lin2023smaug} introduces sparse image patch masks to reduce pre-training costs and improve cross-modal alignment.
EgoVLPv2~\cite{pramanick2023egovlpv2} proposes cross-attention between the backbone encoders to improve both the pre-training efficiency and the downstream task performance.
%Yang et al.~\cite{yang2023learning} models trajectory attention as a two-step attention operation to further emphasize word-related embeddings during the pre-training.

Our method can leverages the pre-trained model during the video-language alignment, which naturally provides reasoning capability with less training cost.

\paragraph{Image-to-Video Transfer Learning}
Due to the high cost of large video dataset collection, many works leverage successful pre-trained image models and transfer the knowledge to video task. Previous works such as ~\cite{arnab2021vivit, bertasius2021space, Heigold_2023_ICCV, Li_2023_ICCV} utilize a pretrained ViT~\cite{dosovitskiy2020image} and aggregate the temporal image feature sequence using transformer block to adapt for video understanding task. For Video-Language task, many works turn to large-scale Vision-Language models as the starting point, such as CLIP~\cite{radford2021learning}, BLIP~\cite{li2022blip}. Many works choose to adapt a pre-trained CLIP model for text-to-video retrieval task, by either augmenting the frame-level spatial features with temporal information~\cite{xue2023clipvip, Deng_2023_ICCV}, or manipulate the cross-modal similarity calculation to get better video-language alignment~\cite{luo2022clip4clip, liu2022ts2net, fang2021clip2video, Wang_2023_ICCV}. Other works ~\cite{NEURIPS2022_a92e9165, Qing_2023_ICCV} focus on parameter efficient fine-tuning for video task by inserting trainable temporal modules into the pre-trained transformer architecture while keeping the rest of model frozen. Recent work SeViLA~\cite{yu2023self} proposes a language-aware frame localizer to sample relevant key frames from videos. %It adopts a pre-trained Vision-Language model BLIP2~\cite{li2023blip} as the frozen backbone and only the adapter QFormer is trained. 
In this paper, we propose a trainable text-guided Frame-Prompter and a QFormer-Distiller module, which help focus more on the important temporal and spatial information from the 2D frames.
These techniques help to efficiently bridge the gap between image-language and video-language learning.

\paragraph{Video Question Answering}
One major downstream task of Video-Language pre-training is Video Question Answering (VQA).
Early works~\cite{amrani2021noise,jiang2020divide,ye2017video} often rely on human annotated datasets, while recent works~\cite{yang2021just,zellers2021merlot,zellers2022merlot} make better use of large-scale data from public.
Along with the quick advances in VQA methods, A lot of benchmark datasets have also been introduced to standardize the model performance comparison, including NExT-QA~\cite{xiao2021next}, STAR~\cite{wu2021star}, How2QA~\cite{li2020hero}, TVQA~\cite{lei2018tvqa}, and VLEP~\cite{lei2020more}.
We benchmark our network mainly on the Video Question and Answering task. 
%It's also the major task to test the effectiveness of the video-language alignment.
%Simple Baselines for Interactive Video Retrieval with Questions and Answers
%Tem-adapter: Adapting Image-Text Pretraining for Video Question Answer
%Discovering Spatio-Temporal Rationales for Video Question Answering
%Knowledge Proxy Intervention for Deconfounded Video Question Answering
%Open-Vocabulary Video Question Answering: A New Benchmark for Evaluating the Generalizability of Video Question

\subsection{Knowledge Distillation}
One of our key component is cross-modal distillation. 
Knowledge Distillation~\cite{hinton2015distilling, zhang2018deep, wang2023triplet} is original proposed for small and efficient models to mimic the softened class distributions, features of large teachers. 
For multi-modalities, researchers explore how to utilize the prior knowledge between different modalities~\cite{gupta2016cross, shen2023auxiliary, shen2023small}. 
On video domain, knowledge distillation has been used for efficient video inference~\cite{mullapudi2019online,khani2021real}, video captioning~\cite{pan2020spatio}, video question answering~\cite{shao2023video}. 
Out of the supervised learning methods mentioned above, the idea of knowledge distillation has also been leveraged in many self-supervised methods for self-supervised video representation learning~\cite{wang2023masked}.

%\TODO{add a section on video frame selection}
\subsection{Frame Selection for Video QA}
Early approaches in Video QA relied heavily on dense sampling methods, where frames are extracted at a fixed interval throughout the video. While straightforward, such methods can lead to excessive computational costs and memory requirements without significantly improving performance. Zhang et al.~\cite{Zhang2018} proposed a more selective strategy, using attention mechanisms to identify key frames that are more likely to contain information relevant to the question. Following this, adaptive frame sampling methods ~\cite{Fan2019, chen2023sas} have gained popularity. These methods aim to dynamically select frames based on the content's relevance to the question, thus optimizing the trade-off between computational efficiency and answer accuracy~\cite{Fan2019}. %Liu et al.~\cite{Liu2021} proposed a method that combines knowledge-enhanced sampling with temporal reasoning, showing that contextual information can guide the selection of informative frames and improve the model's reasoning capabilities

\section{Method}
\label{sec:method}
Our ViLA model tackles the following challenges in large-scale Video-Language learning: how to sample question related frames and how to efficiently transfer video information for pre-trained frozen LLMs. %In this section, we describe our text-guided Frame-Prompter learning module and cross-modal distillation module in detail.

%\paragraph{Problem Statement} Given the input video $X_v$ and question $X_q$, choose the best answer from given set.

\subsection{Model Architecture}
As illustrated in Fig.~\ref{fig:template}, ViLA has four components: a pre-trained frozen large scale visual encoder $E_v$, a Frame-Prompter $F_p$, a QFormer-Distiller $Q_d$  (Querying Alignment Transformer~\cite{li2022blip,li2023blip,dai2023instructblip} with distillation) to fuse and extract text-conditioned visual information, and a pre-trained frozen LLM. 
%\paragraph{End-to-End Training}
We train our Frame-Prompter, QFormer-Distiller and two other frozen modules end-to-end. 
Our training objective includes a distillation loss $L_{\text{distill}}$ and a task loss $L_{\text{vqa}}$.
%\begin{equation}
%L=L_{distill} + L_{task}.
%\end{equation}
%In addition to the distillation loss, the task's loss also has constraints on the selector.
For the multi-choice Video Question Answer task, the task loss is the cross-entropy. 
%\begin{equation}
%L_{task}=\operatorname{Cross-Entropy}(\operatorname{LLM}(Q(E_v(P\cdot X))), GT),
%\end{equation}
%where Ground truth is denoted as $GT$. Therefore, the final optimization objective $L$:

\subsection{Text-guided Frame-Prompter Learning}
\begin{figure}[h]
    \centering
    %\vspace*{-2em}
    \includegraphics[width=\linewidth]{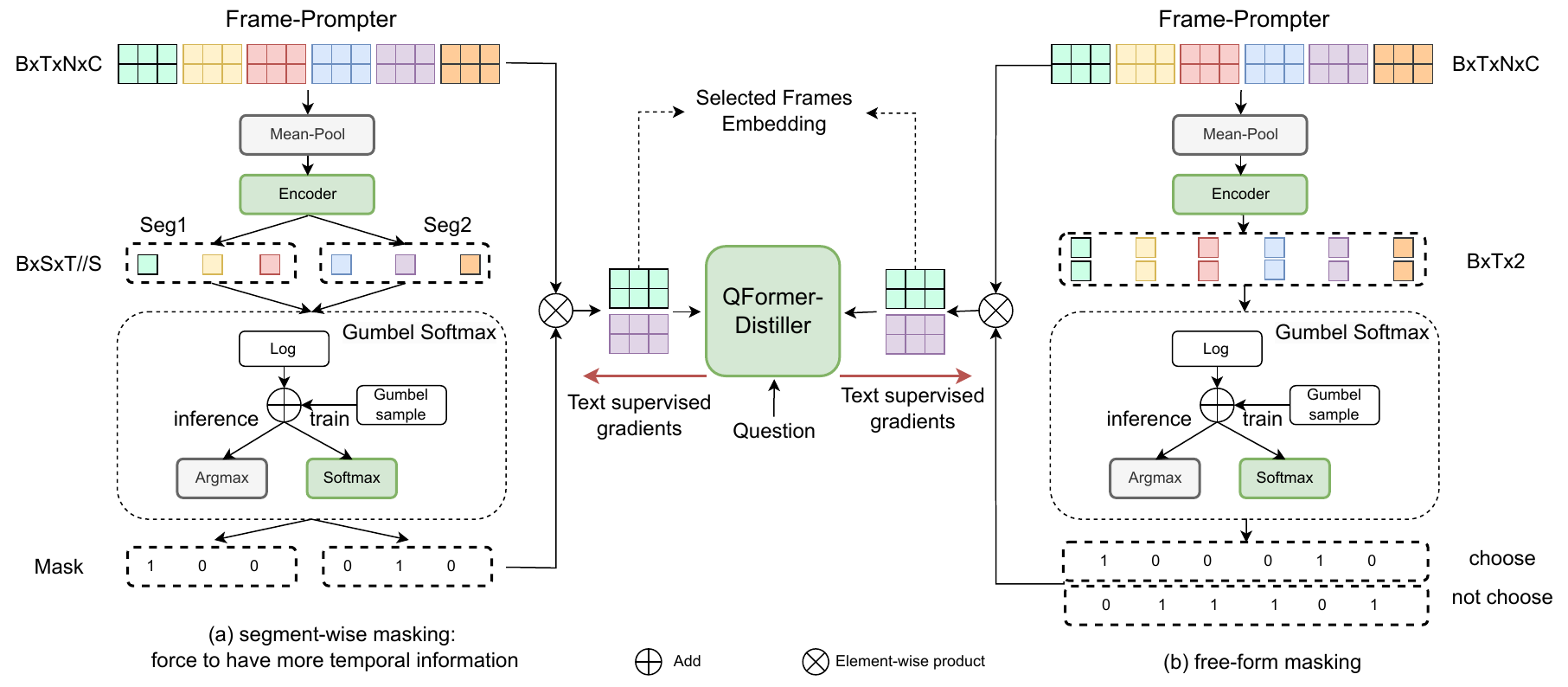}
    \vspace{-6mm}
    \caption{{\bf Text-guided Frame-Prompter.} Here we show the details of our learnable text-guided Frame-Prompter. We design a learnable Frame-Prompter to sample the most text query-related frames, with two design choises (a and b). 
    We choose design (a) for diversified temporal sampling.
    We first encode the mean-pooled segment features.
    We then apply the Gumbel Softmax to compute the segment mask to guarantee the differentiability.
    The selected frames embedding then goes through the QFormer-Distiller. 
    Here $B$ means batch size, $T$ means frame number, $N\times C$ means the frame feature sequences. 
    The Frame-Prompter is learned with the text-supervised gradient. When VQA loss is applied, the input question text-related gradient further flows to the Frame-Prompter. The question text-related gradient guides the Frame-Prompter to select the most critical frames.
    %The question text is encoded and fused with visual embedding in the QFormer. The VQA loss supervise both the QFormer and the Frame-Prompter. 
}
    \label{fig:frame_prompter}
    \vspace{-4mm}
\end{figure}
For the four-dimensional video data, it's impractical and non-efficient to input all frames into a visual encoder model.
%For efficiency consideration and there is lots of redundant information too. 
%And the widely used uniform/random sampling methods don't have any question, which may lose important frames for VQA task and sample frames that are unimportant/irrelevant to language queries when using fewer frames. 
Here we design a text-guided Frame-Prompter for efficient and effective frame sampling.
It is designed to learn attending to fewer but more question-related frames.
%which can sample language queries related frames in a teacher-student manner. 
We start from the uniformly sampled video frames $\{f_1, \dots, f_T\}$, $T$ is frame number. 
%We set is as $32$ frame sampling~\cite{yu2023self} for fair comparison.
As shown in Fig.~\ref{fig:frame_prompter}, these raw frames first go through the pre-trained visual encoder $E_v$,
\begin{equation}
X = \{ x_i | x_i = E_v(f_i), i \in [1,T] \},
\vspace{-2mm}
\end{equation}
where $x_i$ is the visual feature extracted from raw frames and $T$ is the number of frames.
%with shape $B\times T\times N\times C$, where
%$B$ is the batch size, $T$ is the temporal frames number, $N$  is the patch size, and $C$ is the feature dimension. 
We perform mean-pooling on the channels per-patch. The Frame-Prompter input shape is $(B, T, N, C)$, B batch size, T temporal, $(N, C)$ as the frame feature dimension. Mean-pool averages over the $C$ channel. After mean-pool, the feature dimension becomes $(B, T, N)$. Then we divide the feature into S segments $(B, S, T//S\cdot N)$, with $(T//S\cdot N)$ as the segment feature dimension. Then the segmented features goes through FC layer to project to $(B, S, T//S)$ dimension, ready for the gumbel-softmax computation. For free-form frame sampling Fig.~\ref{fig:frame_prompter}(b), after mean-pool, we use FC layer to transfer  $(B, T, N)$ to  $(B, T, 2)$.
Then we encode the mean-pooled visual features onto an embedding using a fully-connected (FC) layer with a layer normalization (LN),
%Then we perform convolution to transform the dimension for frame selection,
\begin{equation}
%\hat{x_i}=W_2*\operatorname{ReLU}(\mathrm{LN}(W_1 * \operatorname{Mean}(x_i)))
\hat{x_i}=W_2*\operatorname{ReLU}(\mathrm{LN}(W_1 * \operatorname{Mean}(x_i))),
\vspace{-2mm}
\end{equation}
where $W_1$, $W_2$ are the learnable weights and $\operatorname{LN}$ is the layer normalization. 
%We perform ablation study on this design choice (Sec.~\ref{sec:fp_encoder}) and it shows the FC with a LN works the best.
%\SY{add a summary for this operation, before diving into details, not sure if we should move some of these to implementation details}
%After $\operatorname{Mean}$ operation, the feature shape is $B\times T\times N$. We reshape the feature into $B\times S \times T/SN$, $S$ is the segment number that divides the video into several segments, we pick one frame in one segment. 
%After convolution $W_1$, the feature shape is $B\times S\times T/S$. Then the feature will go through Layer Normal layer and ReLU. Convolution $W_2$ generates the logits $\hat{x_i}$ for frames in a segment which denotes which frame to select.

Our Frame-Prompter is a differentiable frame selecter. This allows the text-supervised gradients from the QFormer to guide the learning our frame selecter via backpropagation.
The differentiability is achieved through the Gumbel-Softmax~\cite{jang2016categorical}.
%To make this selection process learnable, we need to guarantee all the operations are differentiable. 
%Therefore, we apply  to do the frame selection. 
We have two choices (shown in Fig.~\ref{fig:frame_prompter} a and b) for frame selection before the Frame-Prompter: a) strategic sampling (like: segmented uniform sampling) and b) free-form sampling. 
We choose option (a) to force diverse temporal sampling.
We first convert a segment of concatenated frames feature $s_k=[\hat{x_{i}},\dots,\hat{x}_{T//S}]$ into a categorical distribution $\pi$ through the softmax operation,
\begin{equation}
\pi=\left\{p_k \mid \frac{\exp (s_i)}{ \sum_{j=1}^{S} \exp (s_j)}, k \in [1, T/S], i, j \in [1, S]\right\},
\vspace{-2mm}
\end{equation}
where $S$ is the frame number of one segment.
Then we compute our segment mask $M_S$ using the segment probability $p_k$ and the $g_k$ is sampled from the $\text{Gumbel}(0, 1)$ distribution,
\begin{equation}
M_k=\mathcal{H}\left(\arg \max\left[g_k+\log p_k\right]\right) k\in [1, T/S],
\vspace{-2mm}
\end{equation}
where $\mathcal{H}$ is the one-hot encoding operation. 
During training, we replace the argmax with the softmax for differentiability,
\begin{equation}
%\mathcal{H}\left(\frac{\exp ((\log p_k+g_k) / \tau)}{\sum_{j=1}^{T/S} \exp ((\log p_j+g_j) / \tau)} \right), k, j \in [1, T/S],
\frac{\exp ((\log p_i+g_i) / \tau)}{\sum_{j=1}^{S} \exp ((\log p_j+g_j) / \tau)}, i, j \in [1, S],
\vspace{-2mm}
\end{equation}
where $\tau$ is a tunable temperature. Full mask is $M=\{M_i, \cdots, M_{T//S}\}$. We apply mask $M$ to input frames and then conduct CrossAttention to obtain text guidance by
\begin{equation}
X_{LLM}=\operatorname{CrossAttn}(X\cdot M, X_t),
\end{equation}
where $X_t$ is the text information. This CrossAttention can help the Frame-Prompter learn to choose rich text frames. Finally, we use the task loss below to supervise the learning and choose the frame selection to answer our question: 
\begin{equation}
L_{\text{vqa}}=\operatorname{MSE}(LLM(X_{LLM},X_t), QA_{\text{answer}}) .
\vspace{-2mm}
\end{equation}
$$$$
%We first generate a categorical distribution by using Softmax on each frame feature $x_i$,
%\begin{equation}
%\hat{\pi}=\left\{p_i \mid p_i=\frac{\exp (x_i)}{ \sum_{j=1}^{T/S} \exp (x_j)}, i, j \in [1, T/S]\right\}.
%\end{equation}
%To obtain the frame mask, we draw samples from the categorical distribution with class probabilities $\pi$,
%\begin{equation}
%M=\operatorname{ one\_hot }\left(\underset{i}{\arg \max }\left[g_i+\log \pi_i\right]\right)
%\end{equation}
%where $g_i$ are samples drawn from Gumbel(0, 1). 
%$g_i = -log(-log Gi )$ and $G_j$ is sampled from uniform distribution at range (0,1). 
%To remove the non-differtialble operation $\arg \max$, we use the softmax function as a differentiable approximation for backpropagation:
%\begin{equation}
%\hat{p_i}=\frac{\exp ((\log p_i+g_i) / \tau)}{\sum_{j=1}^{T/S} \exp ((\log p_j+g_j) / \tau)} , i, j \in [1, T/S]
%\end{equation}
%where $\tau$ is the temperature hyperparameter.
%\begin{equation}
%\hat{\pi}=\left\{p_i \mid p_i=\frac{\exp (x_i)}{ \sum_{j=1}^{T/S} \exp %(x_j)}, i, j \in [1, T/S]\right\}.
%\end{equation}
%\begin{equation}
% \pi=\left\{p_i \mid p_i=\frac{\exp (x_i)}{ \sum_{j=1}^{T/S} \exp (x_j)}, i, j \in [1, T/S]\right\},
%\end{equation}
\subsection{Cross-Modal Distillation}
%In this section, we explore the cross-modal , specially for how to transform visual information to the right format which is friendly for LLMs inputs. We propose a TVT (Text-Visual-Text) pattern comprising of Question (Text), Aligned Question-related Visual Feature (visual: feature from QA-Former), Visual-2-Text (Text: key words).
%In this section, we present our QFormer-Distiller,
In this section, we present in detail our cross-modal distillation module,
%designed to further enforce the text conditioned visual information extraction. 
designed to selectively transfer video information ready for LLM consumption.
Video-language alignment performance, unlike image-language alignment, is closely related the selected frames.
%\subsubsection{QFormer-Distiller}
%Language and video alignment is very important for video question answering task. 
Meanwhile, to leverage powerful pre-trained LLMs,
we need to transform the selected visual information to the LLM's input domain. 
We adopt the QFormer proposed in the BLIP models~\cite{li2022blip,li2023blip,dai2023instructblip} for the cross-modal transformation learning.
%BLIP2~\cite{li2023blip} proposed Q-former which is first pretrained with the frozen image encoder for vision-language representation learning and then adapted the output of Q-Former as soft visual prompts for text generation with a frozen LLM. 
%InstructBLIP~\cite{dai2023instructblip} adds task-related question text tokens as additional input to encourage the extraction of task-relevant image features. 

The original QFormer is designed for image-text fusion, and we add a distillation module to make it efficient for video-text fusion.
%We go a step further to design a QFormer-Distiller ($Q_d$).\CK{What do we mean by "going step further" here?}
Like the traditional distillation, our QFormer-Distiller includes a teacher-QFormer and a student-QFormer.
%When training the student QFormer, which applies question text as additional input with a student-teacher leaning paradigm to extract question-relevant video features.  
%Our teacher has a wider receptive field than student, by learning form the teacher's feature and collaborating with frame prompter, student can better model the temporal information with few frames.
We train the teacher-QFormer first (without the student-QFormer) on a wider receptive field.
%Then we train our student-QFormer to narrow down the most important frames.
The student-teacher learning mechanism further encourages both the QFormer and the Frame-Prompter to learn to attend to the most relevant visual information given the input question text. 
We study how QFormer-Distiller affects Frame-Prompter (Sec. ~\ref{sec:fp_qfd}). 
And we find our FP+QFD 4-frames(Acc.{\bf 74.4\%}) model's key-frame selection overlaps {\bf 92.3\%} with the key-frame selected by the FP 8-frames(Acc.{\bf 74.1\%}) model.

Specifically, during student learning process, both the teacher-QFormer and the student-QFormer will take in the video $X_v$ and question text $X_q$.
The concatenated tokenized video and question text will go through the QFormer cross-attention layers:
%will for teacher video input $X$ and question input $X_t$, we first combine question input $X_t$ with query $X_q$:
%\begin{equation}
%\operatorname{Query}=\operatorname{Cat}(\operatorname{E_t}(x_t), X_q)),
%\end{equation}
%then query feature $\operatorname{Query}$ and video feature $E_v(X)$ will go through self-attention $\operatorname{Self\_A}$ and cross-attention $\operatorname{Cross\_A}$:
\begin{equation}
X'=\operatorname{CrossAttn}(E_t(X_q),\operatorname{SelfAttn}(E_v(X_v),\mathbf{q})),
\vspace{-2mm}
\end{equation}
%$\operatorname{FF}(\cdot)$ is feed forward layers. For student video input $P\cdot X$,
%\begin{equation}
%X_q^s=\operatorname{FF}(\operatorname{Cross\_A}(\operatorname{Self\_A}(Query,E_v(P\cdot X)))).
%X'_{\text{student}}=\operatorname{CrossAttn}(E_t(X_q),\operatorname{SelfAttn}E_v(X_v),\mathbf{q})),
%\end{equation}
where $E_t$ is a text tokenizer, $E_v$ is the visual tokenizer, $\mathbf{q}$ is the learnable query token.
We utilize a decoder $D$ to transform the student-QFormer output, ensuring consistency and recoverability to the teacher's feature. This supervision bolsters the effectiveness and performance of the model with fewer frames. Then the distillation objective is defined by:
\begin{equation}
L_{\text{distill}}=\operatorname{MSE}(D(X'_{\text{student}}), X'_{\text{teacher}}).
\vspace{-2mm}
\end{equation}
We carefully design the decoder to be a simple Fully Connected (FC) layer with a layer normalization. 
We show through an ablation study (Sec.~\ref{sec:qfd_decoder}) that this combination works the best. 
Meanwhile, our FP+QFD 4-frames boosts accuracy by {\bf 1.8\%} compared with FP 4-frames model and the key-frame selected overlaps 56.8\% with that of the FP 4-frames model. 
This helps verify our QFD's ability to both boost performance and enhance the key-frame selection.

% \CK{Maybe we can move this paragraph to the beginning of the section, where we motivate why we're doing distilation.}
%And we design a student-teacher pattern for the training to make the student prompter chooses the most informative frames by optimization objective:
%\begin{equation}
%L_{F_p}=\operatorname{MSE}(Q(E_v(P\cdot X)), Q(E_v(X))),
%\end{equation}
%where $Q$ is QA-Former, $\operatorname{MSE}$ is mean squared error loss. 
%\subsubsection{Cross-modal Distillation}
%To solve the problem of efficient training and inference, we propose a new cross-modal distiller in our ViLA network. In our design, we input visual and text information to both teacher and student, this will allow teacher help student to select most related frames, and further extract better question related visual feature. \CK{this paragraph seems out of place. Is it meant to be placed in beginning of the section?}

% \subsubsection{Text-Visual-Text Fusion}
% For LLMs, the best input should be in text format. Most existing works feed LLMs by representing videos using continuous feature vectors or discrete text tokens. In our frame work, we combine them together to grantee most valid information has been input to LLMs. We input discrete text tokens coupled with a pretrained contrastive text model to represent the video information in a text format. 
%\TODO{add a fig for different molding comparison}

\section{Experiments}
%Here we introduce our implementation details, benchmark dataset and the baseline methods.
\subsection{Implementation Settings}
For training, we conduct experiments with 8 × 40GB A100 GPUs. All the models in our experiments are trained using AdamW with cosine learning scheduler. For all the experiments, we use ViT-G(1B)~\cite{fang2023eva} as the visual encoder and Flan-T5 XL (3B)~\cite{chung2022scaling} as large language model. For all the datasets, we use accuracy of choosing the right answer as the metric, and test on the validation dataset. For the temperature $\tau$ in the Frame-Prompter, it changes from 1 to 0.01 by $(0.01)^{(\text{current-step/total-training-step})}$ and is set as 0.01 during testing. And all inference (Infer.) time is evaluated on a single A100 GPU with batch size 4. Please refer supplementary for more training details. 

\subsubsection{Benchmark:}
To demonstrate the effectiveness of our proposed method on the video QA task,
we compare our algorithm with the state-of-the-art (SOTA) methods on 5 datasets including 1 on video event prediction: 
1) \textbf{NExT-QA}~\cite{xiao2021next}, a multi-choice VideoQA benchmark,
%is a benchmark for causal and temporal reasoning  in terms of multi-choice VideoQA. It has different kinds of questions:
with 3 types of questions: Causal (Why, How), Temporal (Previous/Next, Present), and Description (Binary, Location, Count and Other). 
 %It contains a total of 5.4K videos with an average length of 44s and approximately 52K questions. 
2) \textbf{STAR}\cite{wu2021star}, a multi-choice VideoQA benchmark for Situated Reasoning. 
% which contains  video clips with an average length of 12s along with 60K questions. 
It has four kinds of questions: Interaction, Sequence, Prediction, and Feasibility.
3) \textbf{How2QA}~\cite{li2020hero}, a multi-choice VideoQA benchmark with 44k QA pairs for 22k 60-second clips selected from 9035 videos. 
%It provides the start and end points for the relevant moment for each question.
4) \textbf{TVQA}~\cite{lei2018tvqa} is a large-scale video QA dataset 
%based on 6 popular TV shows (Friends, The Big Bang Theory, How I Met Your Mother, House M.D., Grey's Anatomy, Castle).  
with 152K questions along with 21k video clips from 460 hours of video.
%It also provides the start and end points for the relevant moment for each question. 
5) \textbf{VLEP}~\cite{lei2020more} is a video event prediction benchmark,
%that requires the model to predict two future events based on the video premise.  
with 28,726 future event prediction cases.
%from 10,234 diverse TV Shows and YouTube Lifestyle Vlog video clips. 
Following SeViLA~\cite{yu2023self}, we formulate this task as a multi-choice Question Answering.
\begin{table*}[th]
\vspace{-3mm}
\centering
\resizebox{1.0\textwidth}{!}{
\begin{tabular}{cccccccc}
\toprule
Method & \# Frames & Temporal & Causal & Description & Average & \makecell{Intro. \\Param.} & \makecell{Infer. Time \\ (ms/video)}  \\
\midrule 
Just Ask~\cite{yang2021just}  (ICCV2021) & 20 & 51.4 & 49.6 & 63.1 & 52.3 & - & -  \\
All-in-One~\cite{wang2023all}  (CVPR2023) & 32  & 48.6 & 48.0 & 63.2 &  50.6 & - & -\\
MIST~\cite{gao2023mist}  (CVPR2023)  & 32 &  56.6 & 54.6 &  66.9 &  57.1 & - & -\\
HiTeA~\cite{ye2022hitea}  (ICCV2023)  & 16 & 58.3  & 62.4 &  75.6 & 63.1 & -  & -\\
InternVideo~\cite{wang2022internvideo}  (Dec 2022) & 8 &  58.5 & 62.5 &  75.8 &  63.2 & -  & -\\
\textbf{ViLA (Ours)}  & \textbf{1} &   66.5 & 69.3 & 78.0  &  70.5 & 188M  & 64\\
BLIP-2~\cite{li2023blip} (ICML2023) & 4 & 67.2  & 70.3 &  79.8 &  71.5 & - & -\\
\textbf{ViLA (Ours)}  & \textbf{2} &   70.2 & 71.6 & 79.4  &  72.8 & 188M  & 72 \\
SeViLA~\cite{yu2023self}  (NeurIPS2023) & 4 &  \fcolorbox{brown}{white}{\underline{67.7}}  & \fcolorbox{brown}{white}{\underline{72.1}} & \fcolorbox{brown}{white}{\underline{82.2}}  &  \fcolorbox{brown}{white}{\underline{73.4}} & 376M & 301\\
SeViLA~\cite{yu2023self}  (NeurIPS2023)  & 8 &  \fcolorbox{blue}{white}{\underline{67.0}}  & \fcolorbox{blue}{white}{\underline{73.8}} &  \fcolorbox{blue}{white}{\textbf{81.8}} &  \fcolorbox{blue}{white}{\underline{73.8}} & 376M  & 306\\
\midrule
%ViLA (4) (Ours)  &   69.1 & 73.6 & 81.3  &  \textbf{74.2} \\
\textbf{ViLA (Ours)}  & 4 (8 to 4) &   \fcolorbox{brown}{white}{\textbf{71.0}} & 72.9 & \fcolorbox{brown}{white}{\textbf{82.7}}  &  74.3 & 188M  & \textbf{99 (3.04$\times\uparrow$)}\\
\textbf{ViLA (Ours)}  & 4 (32 to 4) &   70.1 & \fcolorbox{brown}{white}{\textbf{73.8}} & 82.1  &  \fcolorbox{brown}{white}{\textbf{74.4}} & 188M  & 208 ($1.45\times\uparrow$)\\
\textbf{ViLA (Ours)} & 8 &   \fcolorbox{blue}{white}{\textbf{71.4}} & \fcolorbox{blue}{white}{\textbf{73.6}} &  \fcolorbox{blue}{white}{\underline{81.4}} & \fcolorbox{blue}{white}{\textbf{74.8}} & 188M & 227 ($1.35\times\uparrow$)\\
%ViLA (Ours)  & 16 &  69.5 & 74.0 &  81.7 &  \textbf{75.0} & 188M \\
\midrule
%ViLA (Ours)  & 32 &  \textbf{72.3}  & \textbf{74.9} &  82.1 &  \textbf{75.6} & 188M & -\\
\textbf{ViLA+LoRA (Ours)}  & \textbf{4} &  71.4  & 74.5 &  80.3 &  75.1 & 188M & 208\\
\textbf{ViLA (Ours)}  & 32 &  \textbf{71.8}  & \textbf{75.3} &  82.1 &  \textbf{75.6} & 188M & -\\
\bottomrule 
\end{tabular}}
\caption{{\bf Comparison Results on NExT-QA dataset.} Here we measure the accuracy of choosing the right answer. 
%Our ViLA outperforms the SOTA method by {\bf 0.9-1.0\%} at 4 frames with {\bf1.45-3.04$\times$} speed up and  by {\bf1.0\%} at 8 frames with {\bf1.35$\times$} speed up.
%and push the accuracy to reach 75.5\% on this dataset at 32 frames setting. 
%And our proposed ViLA with 4 frames even achieved better performance than SeViLA with 8 frames. 
Especially on Temporal and Causal type of questions, our ViLA (using only 4 frames) improves {\bf 3.3\%} and {\bf 1.7\%} respectively, compared with SeViLA. 
We use \textbf{bold-face} font to indicate the best results and \underline{underline} on the second best using the same number of frames (\fcolorbox{brown}{white}{brown box} for 4 frames and \fcolorbox{blue}{white}{blue box} for 8 frames).
ViLA using 2-frames only out-performs BLIP-2 using 4-frames by \textbf{1.3\%}.  ViLA also achieves upto {\bf 3.04$\times$} speedup. It needs to be noted that our ViLA achieves 75.1\% average accuracy with only 4 frames when we finetune LLM with LoRA~\cite{hu2021lora}.
%The comparison with SeViLA shows the effectiveness of our Frame-Prompter and QFormer-Distillation.
%\ljb{can you have the brown box for the text 'brown box' and the same for blue box as well?}
}
\label{tab:nextqa}
\vspace{-10mm}
\end{table*}

%\subsubsection{Baselines}
\subsubsection{Baselines:}
%In our comprehensive evaluation process, 
We compare the performance of our ViLA model with 
several recent prominent models in the field.
These include SeViLA~\cite{yu2023self}, BLIP-2~\cite{li2023blip}, and InternVideo~\cite{wang2022internvideo}, within the context of fine-tuning scenarios. 
For fair comparison with SeViLA and BLIP-2, we employ the Vision Transformer-Global (ViT-G) as the visual encoder and Flan-T5-XL models as the Large Language Model.
%We follow SeViLA to adapt BLIP-2 for video-language training. 
%, respectively, mirroring the configuration that SeViLA has established. 
%This approach ensures consistency and comparability in our assessment.

%Furthermore, adhering to the guidelines set forth in ~\cite{yu2023self}, we have adapted BLIP-2 for video input integration. This adaptation involves a strategic process where we concatenate the visual features extracted from the Q-former, an integral part of the BLIP-2 architecture, and then input these concatenated features into the Flan-T5-XL model. This methodical adaptation allows BLIP-2 to process video inputs effectively, making it a more appropriate candidate for a direct comparison with SeViLA and our ViLA model in these fine-tuning scenarios. \CK{I'm unclear about the reason that this adaptation makes it more appropriate for comparison. Is it because this is what SeViLA did? If so, I think we can just directly say that (and maybe combine with the 2nd half of the previous paragraph.).} %Through these meticulous evaluation methods, we aim to establish a clear understanding of the strengths and weaknesses of ViLA relative to its contemporaries in the field.%
\subsection{Results}
Here we show both quantitative and qualitative (Sec.~\ref{sec:qualitative}) comparison results between our ViLA and SOTA methods on Video QA and Video Event Prediction task (Sec.~\ref{sec:videoqa}). 
Together we present an in-depth analysis on the results. 
Then we conduct ablation study (Sec.~\ref{sec:ablation}) on our proposed Frame-Prompter and QFormer-Distiller module and the design choices within each module.
\subsubsection{Comparison Results on Video QA and Video Event Prediction:}
\label{sec:videoqa}
Overall, we demonstrate that the \textbf{cross-modal key-frame selection matters}.
Our ViLA model out-performs strong SOTA methods across the 4 Video QA benchmark datasets and 1 Video Event Prediction while keeping the inference latency low. 
Especially, we highlight that our models' performance stands out on temporal (including Causal, Interaction, Prediction) type of questions, NExT-QA Temporal (+3.3\%, $3.04\times$ speed up), NExT-QA Causal(+1.7\%, $1.45\times$ speed up), STAR Interaction (\textbf{+4.6\%}, $3.04\times$ speed up), STAR Prediction (+2.8\%, $3.04\times$ speed up), How2QA (+0.3\%, $3.04\times$ speed up), VLEP (+0.7\% with $1.45\times$ speed up, +0.3\% with $4.18\times$ speed up) and TVQA (+1.8\%, $3.04\times$ speed up). 
%Our model also significantly reduced the inference latency by \textbf{67.1\%} compared with SeViLA~\cite{yu2023self} on NExT-QA using 4 frames. 
%The improvement is limited for long videos, How2QA (+0.3\%, $3.04\times$ speed up), due to the limited number of frame ($T=32$) seen by the teacher QFormer.

On the NExT-QA~\cite{xiao2021next} dataset, compared with the SOTA SeViLA~\cite{yu2023self} on 4-frame and 8-frame setting, our ViLA improves by 1.0\% on average accuracy while achieving a ${\bf 3.04}\times$ speed up (see Table~\ref{tab:nextqa}).
%On the NExT-QA dataset, our proposed ViLA reaches 74.4\% at the 4-frame setting, a 1.0\% improvement in performance compared to SeViLA~\cite{yu2023self}.
%Our ViLA key-frame extraction also grasps temporal information better. 
%When going from 4-frame to 8-frame setting for Temporal type of question, our ViLA increases by 1.3\% but SeViLA's decrease by 0.7\%.
%\input{tables/frame_acc}
%But limited on factual (including Descriptive, Feasibility) type of questions. 
%We first evaluate our algorithms on NEXT-QA dataset.
%As illustrated in Table~\ref{tab:nextqa}, our proposed ViLA reaches 74.4\% at the 4-frame setting, a 1.0\% improvement in performance compared to SeViLA~\cite{yu2023self}.
%When configured to the 32-frame setting, ViLA further achieves 75.5\%.
%It's noteworthy that, even when restricted to just 4 frames, ViLA outperforms SeViLA's performance at the 8-frame setting.
%Additionally, our method exhibits substantial improvements in the Temporal category, underscoring the effectiveness and efficiency of our unique distillation design in contributing to the overall performance.
\begin{table*}[th]
\centering
\vspace{-4mm}
\resizebox{1.0\textwidth}{!}{
\begin{tabular}{ccccccc}
\toprule
Method (Frames Number) & Interaction & Sequence & Prediction & Feasibility & Avg. & \makecell{Infer. Time \\ (ms/video)} \\
\midrule 
Flamingo-9B 4-shot ~\cite{alayrac2022flamingo} (30) (NeurIPS2022) & - & - & - & - & 42.8 &- \\
All-in-One~\cite{wang2023all} (32) (CVPR2023)  & 47.5 & 50.8 & 47.7 &  44.0 &  47.5  &-\\
MIST~\cite{gao2023mist} (32) (CVPR2023)  &  55.5 &  54.2 & 54.2  & 44.4  &  51.1   &-\\
InternVideo~\cite{wang2022internvideo} (8) (Dec 2022)  &   62.7 &  65.6 &  54.9  &  51.9  &  58.7 &-\\
BLIP-2~\cite{li2023blip} (4) (ICML2023)  &  \underline{65.4}  &  69.0  &   59.7  &  54.2  &  62.0 &-\\
\textbf{ViLA (2) (Ours)} &   65.0 & 65.4 & 62.2  & 58.8 &  62.9 & 72\\
SeViLA~\cite{yu2023self} (4) (NeurIPS2023)  &  63.7 &  \textbf{70.4} &  \underline{63.1}  &  \textbf{62.4}  &  \underline{64.9} & 301\\
\midrule
%BLVQA (4) (Ours)  &   69.5 & 70.4 & 63.9  & 60.6 &  \textbf{66.1} \\
%ViLA (4) (Ours)  &   69.3 & 70.0 & 63.9  & 64.3 &  \textbf{66.5} \\
ViLA (4) (Ours)  &   \textbf{70.0} & \textbf{70.4} & \textbf{65.9}  & \underline{62.2} &  \textbf{67.1} & 99 ($3.04\times\uparrow$)\\
%ViLA (8) (Ours)  &   70.6 & 74.1 & 66.3  & 60.6 &  \textbf{67.9} \\
\bottomrule 
\end{tabular}}
\caption{{\bf Comparison Results on STAR Video QA benchmark.} 
For Interaction type of question, our ViLA improved \textbf{4.6\%}.
On average, our ViLA out-performs the SOTA method by 2.2\% when using 4 frames with {\bf3.04$\times$} speed up.
%set- ting and achieved new SOTA at 67.9\% on this dataset. 
% We \textbf{bold} the best results and \underline{underline} the second best.
Note that our ViLA using 2-frames out-performs BLIP-2 using 4-frames.
%We achieved state-of-the-art fro all different types of questions and especially for Interaction.
}
\label{tab:star}
\vspace{-8mm}
\end{table*}
\begin{table}[t]
\vspace{-2mm}
\centering
%\resizebox{1.0\columnwidth}{!}{
\begin{tabular}{ccccc}
\toprule
Method   & Frames Numbers & How2QA &  VLEP & TVQA \\
\midrule 
FrozenBiLM  ~\cite{yang2022zero}  (NeurIPS2022) & 10  & 81.5 & - & 57.5 \\
InternVideo  ~\cite{wang2022internvideo} (Dec 2022)  & 8 & 79.0 &  63.9 &  57.2 \\
BLIP-2 ~\cite{li2023blip} (ICML2023)   & 4 & 82.2 &  67.0  &  54.5 \\
SeViLA  ~\cite{yu2023self} (NeurIPS2023) & 4  & \underline{83.6} &  68.9 &  \underline{61.6} \\
\midrule
ViLA  (Ours)  & \textbf{2}& 82.8 &   \underline{69.2} & 60.6 \\
ViLA  (Ours) & 4  & \textbf{83.9} &   \textbf{69.6} & \textbf{63.4} \\
\bottomrule 
\end{tabular}
%}
\caption{{\bf Comparison Results on How2QA, VLEP, and TVQA Video QA Benchmarks.} ViLA improves performance over SeViLA by \textbf{1.8\%} with \textbf{3.04$\times$} speed up on TVQA dataset, 0.7\% with $1.45\times$ speed up on VLEP dataset, and 0.3\% with $3.04\times$ speed up on How2QA dataset at 4 frames setting. Ours 2-frames out-perform SeViLA 4-frames on VLEP by 0.3\% with  ${\bf 4.2}\times$ speed up.} 
\label{tab:vlep_tvqa}
\vspace{-8mm}
\end{table}
%\paragraph{Results on STAR}
%Subsequently, we evaluate our algorithms' performance on the STAR dataset. 

We also test our ViLA model on the STAR~\cite{wu2021star} dataset.
This dataset is challenging due to its diversified type of questions.
%, known for its complexity and diverse range of challenges, serves as an ideal benchmark that highlights the efficacy of our algorithms. 
As shown in Table~\ref{tab:star}, our ViLA model outperforms the several strong SOTA models on average by \textbf{2.2\%} with ${\bf 3.04}\times$ speed up.
Especially on the STAR Interaction and Prediction type of questions, we outperform SeViLA~\cite{yu2023self} by \textbf{4.6\%} and \textbf{2.8\%}. This result further demonstrate the importance of key-frame selection. And our model's advantages in extracting temporal and causal related key-frames.

We further test our model on the longer and larger-scale Video QA benchmark datasets: TVQA~\cite{lei2018tvqa} and How2QA~\cite{li2020hero}, as shown in Table~\ref{tab:vlep_tvqa}.
On the TVQA~\cite{lei2018tvqa} dataset, our ViLA outperforms the SOTA method by \textbf{1.8\%} at the 4-frame setting.
On How2QA, our ViLA improvement is 0.3\% with $3.04\times$ speed up compared with SeViLA~\cite{yu2023self}.
This is partially due to the limited 32-frame teacher-QFormer training. 
On the other hand, compared with Blip-2~\cite{li2023blip}, ViLA outperforms by \textbf{1.7\%}.
This difference again shows the key-frames selected by our ViLA model aligns better with the input question compared with uniform sampling.
To test our algorithm's capabilities, particularly for event prediction, we conduct an additional series of evaluations on VLEP~\cite{lei2020more} dataset. 
%This dataset is specifically designed to benchmark future event prediction. 
At the 4-frame setting, ViLA improves over SeViLA~\cite{yu2023self} by 0.7\% with  $1.45\times$ speed up. it's noteworthy that  ours 2-frames out-perform SeViLA 4-frames on the VLEP dataset by 0.3\% with  ${\bf 4.2}\times$ speed up (Table~\ref{tab:vlep_tvqa}).
% \vspace*{-1em}

\begin{figure*}[tp]
\vspace*{-2em}
    \centering
    \includegraphics[width=\linewidth]{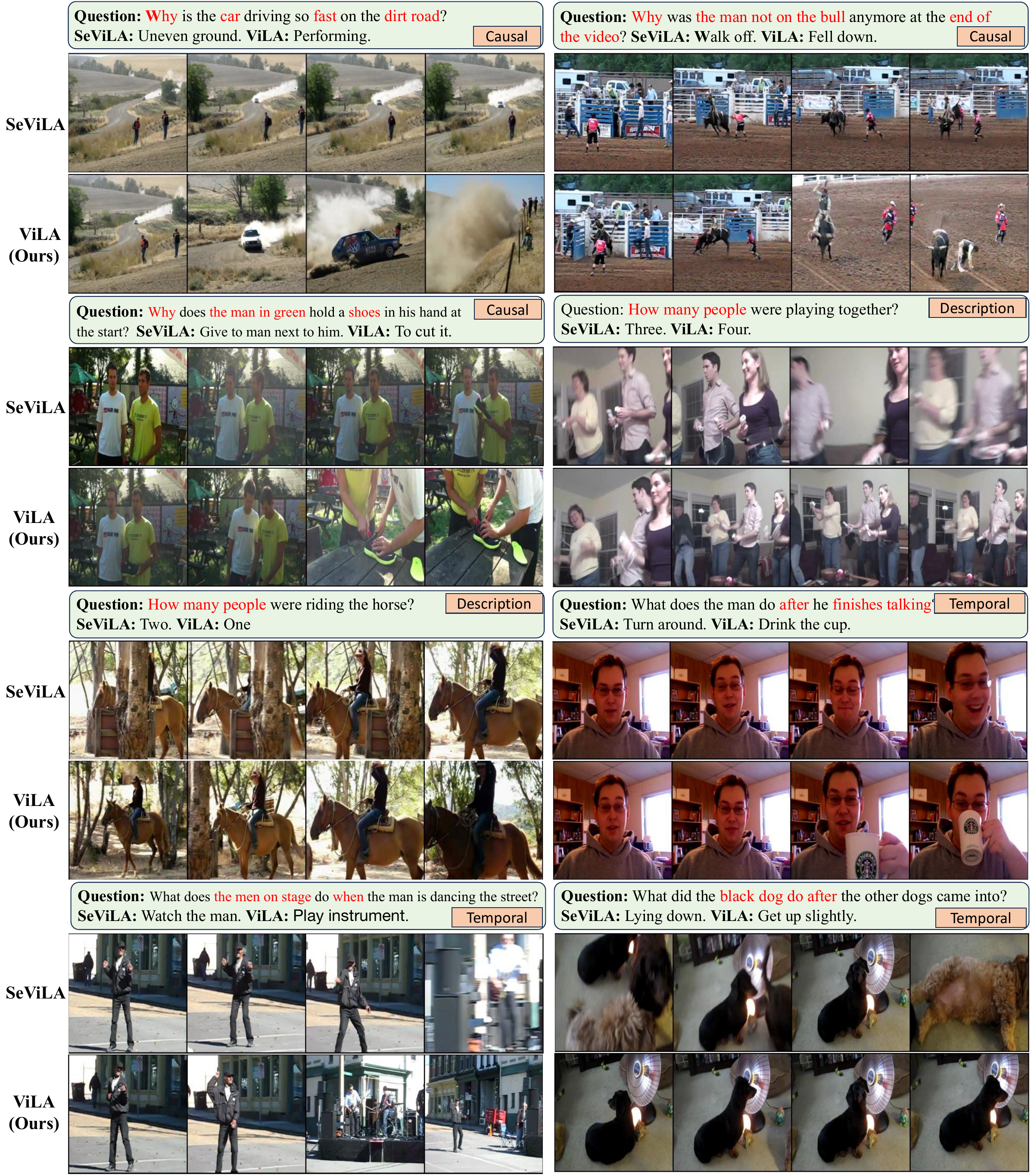}
    \caption{{\bf Key-frame Selection Comparison Results (select 4 frames from 32 frames)}. 
    We compare frames selected by our ViLA compared with that from the SOTA SeViLA~\cite{yu2023self} method.
    Across different type of questions, especially the Causal, Temporal type questions, keyframes selected by our network is more relevant and better related to the question. 
    %(See Supplementary Document for more visualizations.)
    %In the first (col 1 row 1) example in Figure~\ref{fig:visual}, our ViLA locate frames where ``the man sweep the leave with the boy''.
%For Temporal type of questions, our ViLA selects frames that closer to the temporal specified in the question.
%In the 7th (col 1 row 4) and 8th (col 2 row 4) example in Figure~\ref{fig:visual}, we choose key-frames around the dog playing in the pool and the black dog (vs. SeViLA on the brown dog).
    }
    \label{fig:visual}
    \vspace{-2em}
\end{figure*}
\vspace*{-1em}
\subsubsection{Qualitative Results:}
\vspace*{-0.5em}
\label{sec:qualitative}
We qualitatively compare the key-frames selected by our ViLA with the ones from SeViLA~\cite{yu2023self} on different type of questions. 
As shown in Figure~\ref{fig:visual}, our network select more question-related key-frames across different question types (including Causal, Temporal and Description or Factual). 
In the first (col 1 row 1) example in Figure~\ref{fig:visual}, our ViLA locates the frames that visibly show the ``car and the dirt'', but the frames selected by SeViLA focuses on the ``road''.
In the second (col 2 row 1) example, we locate the frames with the ``the man not on the bull at the end'', but the frames selected by SeViLA focuses on the ``man on the bull at the end''.
And in the fourth (col 2 row 2) example, we locate 3 frames with the ``four people'' according to the question, but none of the frames selected by SeViLA shows ``four people''. 
For Temporal type of questions, our ViLA is also capable of selecting frames that are closer to the action specified in the question.
In the seventh (col 1 row 4) and eighth (col 2 row 4) example in Figure~\ref{fig:visual}, we locate key-frames around the ``the men on stage'' and the ``black dog'' (vs. SeViLA has 2 frames focusing on the ``brown dog''). 

\begin{figure*}[tp]
    \centering
    \includegraphics[width=\linewidth]{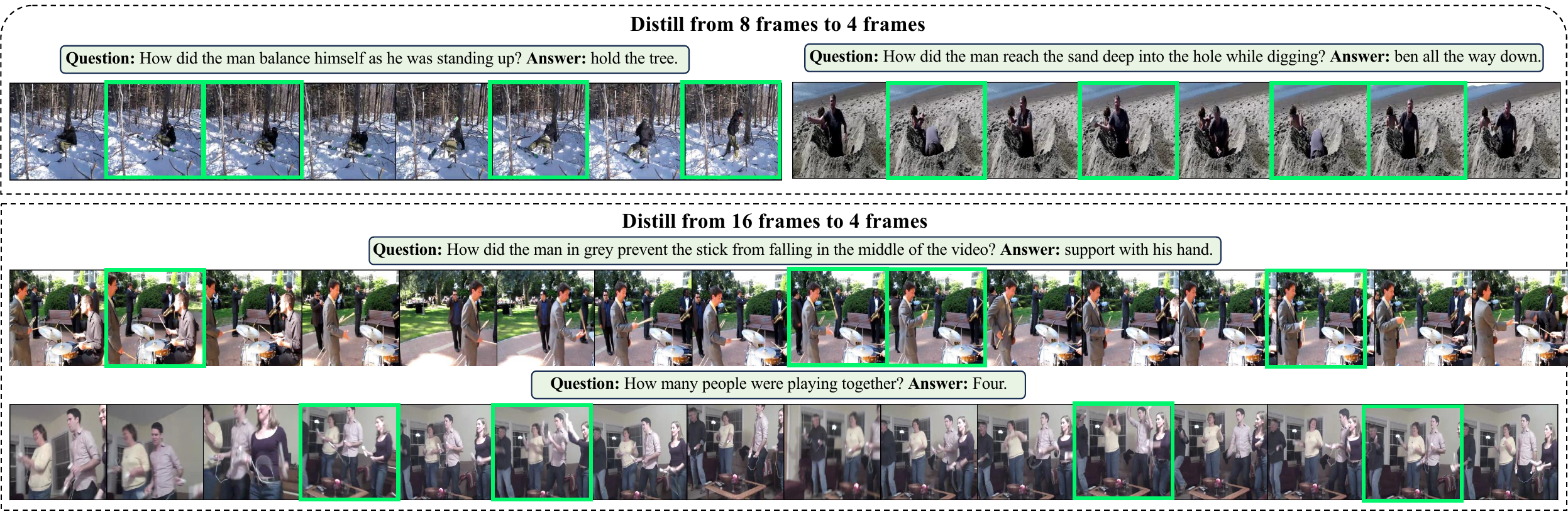}
    \vspace*{-2em}
    \caption{{\bf QFormer-Distiller Results Visualization.} Here we visualize the keyframes selected after cross-modal distillation.
    After distillation, we can select the most question-relevant frames even from 16 frames.
    %(See Supplementary Document for more visualizations.)
    }
    \label{fig:visual_dist}
    %\vspace{-1em}
\end{figure*}

In addition, we qualitatively check the key-frames selected though our QFormer-Distiller.
We show in Figure~\ref{fig:visual_dist} training our QFormer-Distiller helps select the most question related frames.
In the second example, we select frames that shows the ``deep into the whole''. 
And in the third example, out of the 16 frames, we pick up the frames that have both the ``man in grey and the stick falling''.

\subsection{Ablation Study}

\label{sec:ablation}
Here we demonstrate the effectiveness of each proposed module: the text-guided Frame-Prompter and the question-relevant QFormer-Distiller. 
We also validate the decoder design choice within the QFormer-Distiller. 

\subsubsection{Frame-Prompter and QFormer-Distiller Ablation:}
\label{sec:fp_qfd}

\begin{table}[h]
\centering
%\resizebox{1.0\columnwidth}{!}{
\begin{tabular}{c c c c c}
\toprule
Components & STAR & VLEP & TVQA & NExT-QA  \\
\midrule 
base (BLIP-2)  & 62.0 & 67.0 & 54.5 &  71.5 \\
base+QFormer-Distiller  &   64.9 & 68.6 & 62.2  &  73.5 \\
base+Frame-Prompter  &   65.3 & 68.8 & 62.7  &  73.6 \\
base+QFormer-Distiller+Frame-Prompter  &   66.5 & 69.6 & 63.4  & 74.4 \\
\bottomrule 
\end{tabular}
%}
\caption{{\bf Frame-Prompter and QFormer-Distiller Ablation Results.} Across all four VideoQA datasets, we observe that both Text-aware Frame-Prompter and cross-modal QFormer-Distiller contribute significantly to our final performance.
We highlight that on STAR, adding our QFormer-Distiller improves the accuracy by \textbf{2.9\%}. 
Our Frame-Prompter further boost the accuracy by \textbf{1.6\%}.
}
%Cross-Modal Alignment (CMA), Text-aware Frame Prompter (QFP).}
\label{tab:comp}
\vspace{-4mm}
\end{table}
%In our study, we have undertaken a thorough evaluation to ascertain the individual contributions of each component within our methodology.
%Here we study the effectiveness of our proposed Frame-Prompter and QFormer-Distiller. %
%The detailed results of this evaluation are systematically presented in Table~\ref{tab:comp}. 

Here we ablate our new QFormer-Distiller (QFD) and Frame-Prompter (FP).
To summarize, our QFormer-Distiller and our Frame-Prompter each contributes 50\% to the overall performance improvement across the 4 Video QA benchmarks.
Specifically, our QFormer-Distiller boost performance by \textbf{2.9\%} and our Frame-Prompter by \textbf{1.6\%} on the STAR~\cite{wu2021star} dataset, as shown in Table~\ref{tab:comp}. This shows both modules are critical to achieving the desirable performance.

We further explore how QFormer-Distiller (QFD) affect Frame-Prompter (FP). We find our FP+QFD 4-frames(Acc.{\bf 74.4\%}) model's key-frame selection overlaps {\bf 92.3\%} with the key-frame selected by the FP 8-frames(Acc.{\bf 74.1\%}) model.
Meanwhile, our FP+QFD 4-frames boosts accuracy by {\bf 1.8\%} compared with FP 4-frames model and the key-frame selected overlaps 56.8\% with that of the FP 4-frames model. 
This helps verify our QFD's ability to both boost performance and enhance the key-frame selection.

%First, we test on the 4 VideoQA datasets: STAR, VLEP, TVQA and NExT-QA to understand the impact of ablating each component.
%Our analysis encompasses four distinct datasets, providing a comprehensive understanding of the impact each element has on the overall performance.
%As shown in Table~\ref{tab:comp},
%Upon our results, it consistently shows that 
%both the Frame-Prompter (FP) and the QFormer-Distiller (QFD) contribute to the final performance metrics across the 4 Video QA datasets.

%The text-guided Frame-Prompter handles keyframe selection, and the QFormer-Distiller aligns different modalities by using knowledge distillation.%
%This evaluation underscores the significance of both IFP and CMA in our overall methodology.

\subsubsection{QFormer-Distiller Decoder Ablation:}
\label{sec:qfd_decoder}
We study the design choice of the decoder of the QFormer-Distiller.
%Our Frame-Prompter as shown in Figure~\ref{fig:frame_prompter} is a shallow network.
One of the key components is the decoder before computing the distillation loss.
%In our the decoder component of the Frame-Prompter. 
Design choices for this component includes the number of Fully Connected layer (FC) and where to put a layer normalization (LN).
From Table~\ref{tab:frame_prompter}, we show that the simple FC with a LN (after FC) works the best. 
%We experiment with the combination of various basic layers, including the Fully Connected layer (FC) and the Layer Normalization (LN), and show the result in Table~\ref{tab:frame_prompter}.
\begin{table}[t]
\vspace{1em}
\centering
\begin{tabular}{c c c c c}
\toprule
Frame Prompter Decoder & Temporal & Causal & Description & Average  \\
\midrule
FC  &   68.5 & 70.9 & 79.3  &  72.4\\
FC+LN  &    70.1 & 73.8 & 82.1   &  74.4\\
FC+LN+GELU+FC     & 69.7 &  73.1 & 81.6 &  74.1 \\

\bottomrule 
\end{tabular}
\caption{{\bf QFormer-Distiller Decoder Ablation on NExT-QA.} We find that a simple Fully Connected layer (FC) with a Layer Normalization (LN) works  best across 
%It keeps \textbf{efficient and effective}. 
Temporal, Causal, Description. It is efficient and effective. GELU is activation function.}
\label{tab:frame_prompter}
\vspace{-5mm}
\end{table}
%We found that a simple FC layer augmented with a LN layer yields the best performance.
The FC layer provides the necessary computational structure, while the LN layer stabilizes the learning process. This configuration balances the effectiveness and operational efficiency, making it a well-suited choice for our method.

\subsection{More Discussion}
More recent works~\cite{ko2023large, yu2023self} 
have shown powerful networks such as LLMs learns most of the data distribution during the pre-training stage. 
This is one of the major reason why prompt-learning is very effective.
Our work is inspired from the prompt-learning. 
We hope to leverage LLM through proper visual prompting without affecting the generalization ability of the LLM, and this won't affect LLMs' original ability on language tasks. 

However, for the VQA task itself, optimal performance in this task often necessitates training the Language Model (LLM). Therefore, we conduct an ablation study on NExT-QA that fine-tunes the LLM by using LoRA during the training. Our ViLA achieves 75.1\% average accuracy with only 4 frames. This demonstrates that our ViLA has a strong adaptation ability.

Furthermore, to evaluate the generalization of our ViLA, we replace Flan-T5 with llama on NExT-QA, baseline with llama (68.6\%) vs. ViLA with llama ( 72.7\%), which shows that ViLA can adapt to different LLMs.

\section{Conclusion, Limitation and Future Work}
\label{sec:conclusion}
%Pre-trained large vision-language models have shown promising results on problems such as visual question answering. Therefore, most of the state-of-the-art image-language models are built from large pre-trained vision-language models.
%However, videos, unlike images have one additional temporal dimension. For any video understanding tasks, how to efficiently and effectively sample image frames when 
% adapting pre-trained large vision-language model to video-language alignment is still the major challenge. 
%The advances in Large Language Models (LLMs) bring great boost to natural language understandings. New challenge emerges on 
How do we properly ingest visual information to LLMs to utilize its capability effectively in cross-modal tasks?
In this work, we present a new ViLA network with a new text-guided Frame-Prompter to smartly sample important frames, together with a cross-modal temporal distillation (QFormer-Distiller) for efficient and effective video-language alignment.  
From our experiments, our ViLA outperforms SOTA methods on four video question answering benchmarks and one video event prediction benchmark, 
especially on the temporal and interaction type of questions. We demonstrate that cross-modal keyframe selection is key to successful video-language alignment task execution.
Due to resource constraints, we only evaluate on LLMs with the number of parameters not larger than 13 billions.
%We demonstrate that cross-modal keyframe selection is key to successful video-language alignment task execution.
%On the other hand, we find our performance improvement is limited on factual (like Descriptive, Feasibility and etc.) type of questions. 
We plan to continue research on the design of our Frame-Prompter, especially on video-language alignment for long video segments.

\bibliographystyle{splncs04}
\bibliography{main} 

%\pagebreak

%\input{12_appendix}

\end{document}